\documentclass[letterpaper]{article} 
\usepackage{aaai25}  
\usepackage{times}  
\usepackage{helvet}  
\usepackage{courier}  
\usepackage[hyphens]{url}  
\usepackage{graphicx} 
\usepackage{natbib}  
\usepackage{caption} 
\usepackage{algorithm}
\usepackage{algorithmic}
\usepackage{cite}
\usepackage{amsmath,amssymb,amsfonts}
\usepackage{array}
\usepackage{algorithmic}
\usepackage{graphicx}
\usepackage{textcomp}
\usepackage{xcolor}
\usepackage{booktabs}
\usepackage{float}
\usepackage{subcaption}
\usepackage{fancyvrb}
\usepackage{newfloat}
\usepackage{listings}
\DeclareCaptionStyle{ruled}{labelfont=normalfont,labelsep=colon,strut=off} 
\floatstyle{ruled}
\newfloat{listing}{tb}{lst}{}
\floatname{listing}{Listing}
\title{Moonshine: Distilling Game Content Generators into Steerable Generative Models}
\author {
    Yuhe Nie\equalcontrib\textsuperscript{\rm 1},
    Michael Middleton\equalcontrib\textsuperscript{\rm 1},
    Tim Merino\textsuperscript{\rm 1}
    Nidhushan Kanagaraja\textsuperscript{\rm 1}
    Ashutosh Kumar\textsuperscript{\rm 1}\\
    Zhan Zhuang\footnotemark[2]\textsuperscript{\rm 2,3}
    Julian Togelius\thanks{Corresponding authors.}\textsuperscript{\rm 1}
}
\affiliations {
    \textsuperscript{\rm 1}New York University Game Innovation Lab, Brooklyn, New York, USA\\
    \textsuperscript{\rm 2}Southern University of Science and Technology, Shenzhen, China\\
    \textsuperscript{\rm 3}City University of Hong Kong, Kowloon, Hong Kong SAR China\\
    yn2273@nyu.edu, Michael.Middleton@nyu.edu, tm3477@nyu.edu, nk3755@nyu.edu, ak10514@nyu.edu, 12250063@mail.sustech.edu.cn, Julian@Togelius.com
}

\usepackage{bibentry}

\newcommand{\revision}[1]{{\color{black}{#1}}}

\begin{document}

\maketitle
\begin{abstract}

Procedural Content Generation via Machine Learning (PCGML) has enhanced game content creation, yet challenges in controllability and limited training data persist. This study addresses these issues by distilling a constructive PCG algorithm into a controllable PCGML model. We first generate a large amount of content with a constructive algorithm and label it using a Large Language Model (LLM). We use these synthetic labels to condition two PCGML models for content-specific generation, the Five-Dollar Model and the Discrete Diffusion Model. This neural network distillation process ensures that the generation aligns with the original algorithm while introducing controllability through plain text. We define this text-conditioned PCGML as a Text-to-game-Map (T2M) task, offering an alternative to prevalent text-to-image multi-modal tasks. We compare our distilled models with the baseline constructive algorithm. Our analysis of the variety, accuracy, and quality of our generation demonstrates the efficacy of distilling constructive methods into controllable text-conditioned PCGML models.

\end{abstract}



\begin{links}
    \link{Datasets}{huggingface.co/datasets/DolphinNie/dungeon-dataset}
\end{links}

\section{Introduction}

Procedural Content Generation (PCG) is crucial to many types of video games, including roguelikes and other games that are designed around infinite content variety. PCG is also important for empowering users and designers to create various types of game content. Many games feature generators for content such as quests, maps, levels, and items. These generators are built on different algorithms and underlying principles, including search, constraint satisfaction, machine learning, and hand-crafted rules.


An important concern in content generation is controllability. Many content generators feature this to some degree; in its simplest form, controllability can be selecting a map size or number of enemies. While this basic level of control may be suitable for some applications, it can be limiting in others. On the opposite end of the spectrum is text-conditioned content generation. In this form, any idea that can be expressed through natural language can serve as guidance for a generator. The rise of Text-to-Image (T2I) models has demonstrated the value of such flexible means of control over content generation. While Text-to-Image capabilities have improved massively in recent years, these models are not well suited for many game-content generation tasks, where each game has a unique set of constraints.

Procedural Content Generation via Machine Learning (PCGML) \cite{summerville2018procedural} uses machine learning models to \textit{learn} from existing game content and generate diverse artifacts. \revision{However,} a common challenge for PCGML methods is the lack of quality data \revision{for training}. \revision{This challenge is exacerbated when training controllable generators, which require diverse and descriptive labels for the data. To enable text-conditioned PCGML, the demand for large, richly labeled datasets becomes even greater. Data labeling, requiring significant human labor, is often infeasible for most game developers.}

\revision{To address this hurdle, we propose a novel approach that integrates traditional PCG algorithms with the capabilities of Large Language Models (LLMs) to generate large synthetic datasets for text-conditioned PCGML. Specifically, we leverage the PCG algorithm from the game Brogue \cite{brogue} to create a diverse collection of game maps and automatically generate descriptive labels for each, referenced from a small set of human examples. This synthetic dataset is then used to train two Text-to-game-Map (T2M) generative models: the Five-Dollar Model \cite{merino2023five}, which is a feed-forward model, and a Discrete Diffusion Model, which will be explained in detail in this paper. The outputs of these models are evaluated through both quantitative and qualitative metrics. Ultimately, our approach enables generative models to replicate the behavior of a black-box PCG algorithm using only its generated artifacts, while incorporating natural language text conditioning.} 


\revision{This approach can be viewed as a form of knowledge distillation —— traditionally used to transfer the capabilities of one neural network to another. However, instead of distilling between neural networks, we are transferring the behavior of an arbitrary algorithmic process into a neural network. While this adaptation of distillation lacks a formal theoretical foundation, it demonstrates promising results in practice. Inspired by its unconventional and experimental nature, we refer to the overall approach as \textit{Moonshine}.}


In summary, our contributions are:

\begin{enumerate}
  \item We devise a strategy for creating extensive synthetic datasets for text-conditioned PCGML using traditional PCG methods combined with LLMs \revision{and few-shot human labels}.
  \item We frame text-conditioned PCGML as a Text-to-game-Map (T2M) task, analogous to the Text-to-Image (T2I) task, \revision{but operating in a discrete domain}, and propose two training methodologies to address its unique challenges.
  \item \revision{We provide a distillation process to convert any traditional PCG algorithms into controllable text-conditioned PCGML models effectively.}
\end{enumerate}

\section{Related Work}
\subsection{Procedural Dungeon Generation} 

Text-to-game-Map (T2M) generation is a form of Procedural Dungeon Generation, \revision{where PCG algorithms are} used to generate maps for e.g. dungeon crawler-type video game \cite{viana_survey_2019}. \citet{togelius2011search} identified two primary approaches in the current landscape of dungeon generation: constructive algorithms and search-based algorithms~\cite{togelius2011search}. Constructive algorithms directly generate dungeon levels through a variety of methods, such as Cellular Automata \revision{\cite{wolfram1983statistical} and} Generative Grammars \revision{\cite{horrocks2014generative}}.
Search-based algorithms utilize metaheuristics, such as evolutionary algorithms \revision{\cite{raffe2014integrated} and Wave Function Collapse~\cite{nie2024nested}} to optimize dungeon content against a set of criteria, using cycles of generation, evaluation, and selection to iteratively refine solutions. Human controllable content generation is important for designers to input their own preferences into generation. However, \citet{viana_survey_2019} highlighted the scarce reliance on mixed-initiative approaches within dungeon generation, where human design interacts with computer-generated content.

In the context of PCG, \citet{6661386} proposed a methodology for generating dungeons in role-playing games using a combination of graph theory and geometric algorithms. Their approach focuses on creating diverse and engaging gameplay experiences through the procedural generation of dungeon layouts, room connections, and game objects. The authors demonstrated the effectiveness of their approach through an evaluation of the generated dungeons, showcasing the potential of procedural content generation in enhancing gameplay variety and player engagement. They identified controllability as the key challenge—ensuring that procedurally generated dungeons adhere to the desired criteria for gameplay progression, difficulty curves, and \revision{thematic consistency}.

\subsection{Synthetic Data Labeling}
Most synthetic data labeling approaches come from the domain of text-to-image (T2I) generative models and captioning. \citet{he2024automated} introduced PRISM, an algorithm that automatically identifies human-interpretable and transferable prompts that can effectively generate desired concepts. This is particularly noteworthy as prompt engineering has been recognized as an effective method for controlling the output of T2I generative models, but it is also laborious due to the need for manually crafted prompts.

\revision{Expanding on the theme of automation in \revision{machine learning}, \citet{Yang_2023_ICCV} introduced Adaptive Language-Image Pre-training (ALIP), a bi-path model that incorporates supervision from both raw text and synthetic captions. This approach addresses the challenges posed by intrinsic noise and mismatched image-text pairs in web data, which can negatively impact representation learning. The effectiveness of ALIP was demonstrated through experiments across various model scales and pre-training datasets, achieving state-of-the-art performance on multiple downstream tasks.}


\revision{\subsection{Text-to-Image Generation}
In the field of computer vision, Generative Adversarial Networks (GANs)~\cite{reed2016generative} have been widely used for text-to-image generation, leveraging adversarial training to produce high-quality outputs. However, GANs often struggle with mode collapse, and are not always responsive to conditional inputs. Diffusion Models~\cite{saharia2022photorealistic} address these issues by iteratively denoising random noise to generate images, achieving superior fidelity and diversity. Key components in Diffusion Models include Variational Autoencoders (VAEs)~\cite{kingma2019introduction}, which provide structured latent spaces for controlled generation, and the Contrastive Language-Image Pre-training (CLIP)
~\cite{radford2021learning}, which maps images and text into a shared multimodal embedding space, consists of an image encoder for visual inputs and a text encoder for textual descriptions. These encoders are trained jointly using a contrastive loss, enabling CLIP to learn a wide range of visual concepts directly from natural language descriptions. This allows it to perform various vision tasks in a zero-shot manner by comparing the embeddings of text prompts with those of images to identify the best match.}

\section{Synthetic Dataset Generation}
Our pipeline for generating synthetic \revision{$\{map, description\}$} dataset follows three main steps:
\begin{enumerate}
  \item Extract a map data point from a game using a traditional PCG algorithm.
  \item Perform metadata and heuristic analysis on the map.
  \item Input metadata and heuristics into an LLM prompt to generate a descriptive map label.
\end{enumerate}  

\subsection{Map Extraction}

We extract the map data from the open-source rogue-like dungeon game Brogue \cite{brogue}. This game uses a traditional constructive PCG algorithm to generate varied and complex dungeon maps. Unlike many constructive algorithms such as Binary Space Partitioning or Cellular Automata \cite{shaker2016procedural}, Brogue's algorithm ensures \textit{playability}, \revision{considering the connectivity, terrain layout, and ecosystem} when generating various difficulties for players.

The original Brogue game contains diverse tilesets with terrain, objects, monsters, items, and more. This research focuses on the terrain tiles to simplify the original game content. We scale the maps to a consistent size $32\times32$ \revision{pixels} with a tileset of $14$ terrain tiles, as detailed in Table~\ref{tab:tileset}.

\begin{table}[H]
  \centering
\begin{tabular}{cccccc}
\hline
\textbf{Tile} & \textbf{Desc.}  & \textbf{Tile} & \textbf{Desc.}  & \textbf{Tile} & \textbf{Desc.}
\\\hline
\raisebox{-0.25\height}{\includegraphics[width=0.05\columnwidth]{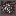}} & Ashes &  
\raisebox{-0.25\height}{\includegraphics[width=0.05\columnwidth]{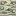}} &  Bog &
\raisebox{-0.25\height}{\includegraphics[width=0.05\columnwidth]{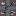}} & Bridge
\\\hline
\raisebox{-0.25\height}{\includegraphics[width=0.05\columnwidth]{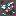}} & Crystal &
\raisebox{-0.25\height}{\includegraphics[width=0.05\columnwidth]{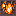}} & Fire &
\raisebox{-0.25\height}{\includegraphics[width=0.05\columnwidth]{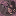}} & Fungus 
\\\hline
\raisebox{-0.25\height}{\includegraphics[width=0.05\columnwidth]{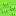}} & Grass &
\raisebox{-0.25\height}{\includegraphics[width=0.05\columnwidth]{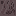}} & Ground &
\raisebox{-0.25\height}{\includegraphics[width=0.05\columnwidth]{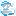}} & Ice 
\\\hline 
\raisebox{-0.25\height}{\includegraphics[width=0.05\columnwidth]{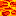}} & Lava &
\raisebox{-0.25\height}{\includegraphics[width=0.05\columnwidth]{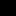}} & None &
\raisebox{-0.25\height}{\includegraphics[width=0.05\columnwidth]{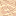}} & Sand
\\ \hline
\raisebox{-0.25\height}{\includegraphics[width=0.05\columnwidth]{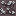}} & Stone &
\raisebox{-0.25\height}{\includegraphics[width=0.05\columnwidth]{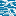}} & Water 
\\\hline
\end{tabular}
\caption{The details of dungeon tileset with descriptions.}
\label{tab:tileset}
\end{table}

We create a dataset of maps from the game, split into 49,000 training points, 14,000 test points, and 7,000 validation points. We open-sourced this dataset and uploaded it to Hugging Face's dataset repository. 

\subsection{Map Metadata Analysis}

\begin{figure}
    \centering
    \includegraphics[width=0.3\textwidth]{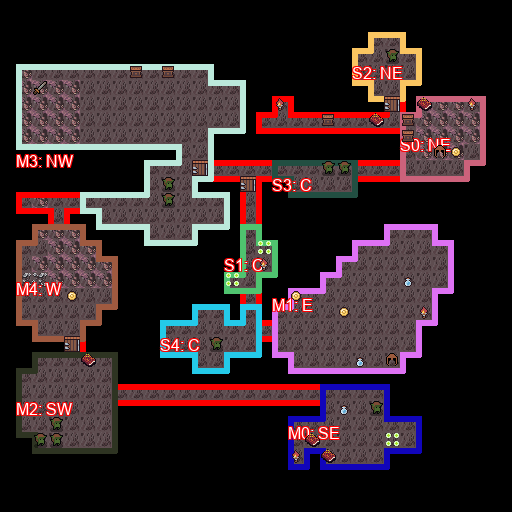}
    \caption{A visualized example of map metadata information. Independent rooms and connecting paths are colored. Room labels and directions are overlayed as text.}
    \label{fig:metadata}
\end{figure}

We perform multiple heuristic calculations to extract metadata for each map. The metadata will then be provided in the LLM's prompt to generate a descriptive label for the map. A visual example of metadata is shown in the Fig. \ref{fig:metadata}.

\begin{enumerate}
    \item We place a binary mask over each independent room and connecting path.
    \item We divide the map into a grid of cardinal and inter-cardinal directions (N, S, E, W, NE, NW, SE, SW), and assign each room a direction using its midpoint.
    \item We extract the tile counts for each room, order them by quantity, and add them to the room's description. 
    \item We record the connected room pairs and the path to connect the pair.
\end{enumerate}

\subsection{Synthetic Description Generation}

\subsubsection{Pre-generation Prompt} 

We structure our pre-generation prompt into four key sections: The \textit{Setting} introduces the task and clarifies the LLM's goal. The \textit{Response} format specifies the exact output structure required. The \textit{Examples} provides a few-shot demonstration of human-authored map descriptions to guide the LLM. Finally, the \textit{Rules} outlines specific constraints that the LLM must follow during generation.

\revision{\subsubsection{Generation Round} 

In each following round of conversation}, we provide the LLM with an integer grid representation of the map, a dictionary of integers map to the tile names, and the meta-data information. We task the LLM to generate 10 text descriptions for each map. This process is repeated for each unique map. 

\section{Text-to-game-Map Generation}

\subsection{Task Definition}
Our approach to Text-to-game-Map (T2M) generation is similar to Text-to-Image generation but differs due to the discrete grid nature of the map data. Rather than generating continuous pixel values, our model generates a grid of probability vectors and then selects the most probable tile at each cell.

\subsubsection{Map Representation} Each map $m$ in our dataset is represented as a three-dimensional matrix of dimensions $H \times W \times C$ where $H$ and $W$ denote the map's height and width, respectively, and $C$ represents the tileset size. Each cell $m_{i,j}$ within the matrix is a vector of length $C$ and follows a classification distribution, ensuring that the sum of probabilities across all tiles in a cell equals one.

In the training dataset, each map is represented in a one-hot encoded format, where each cell is set to indicate a specific tile choice $m_{i, j, k^*} = 1$, \revision{while} all other tiles are set to zero, $m_{i, j, l} = 0$ for $l \neq k^*$. \revision{Mathematically, this is expressed as:}
\begin{equation*}
    \forall i \in [0, H];\  j \in [0, W];\ \sum_{k=1}^C m_{i,j,k} = 1.
\end{equation*}

\subsubsection{Text Embedding Model}

\revision{To extract text embedding vectors, we employ the pre-trained model \textit{gte-large-en-v1.5}~\cite{li2023towards}, which is one of the best-performing small models on the massive text embedding benchmark. This model supports a maximum input size of 8192 tokens and generates 1024-dimensional embedding vectors.}

\subsubsection{Generative Model} We define a comprehensive generative model $\mathcal{G}$, which produces maps $M$ conditioned on text embedding $\mathbf{t}$. This model, parameterized by weights $\mathbf{\theta}$, may optionally incorporate input $\mathbf{z}$ from a noise distribution to enhance generation diversity. By integrating textual context into the map generation process, the model produces content that is both contextually relevant and varied, based on the input text.

We evaluate two multi-modal strategies to investigate the effects of distilling the constructive PCG algorithm into controllable T2M models.

\begin{figure*}[h]
    \centering\includegraphics[width=0.85\textwidth]{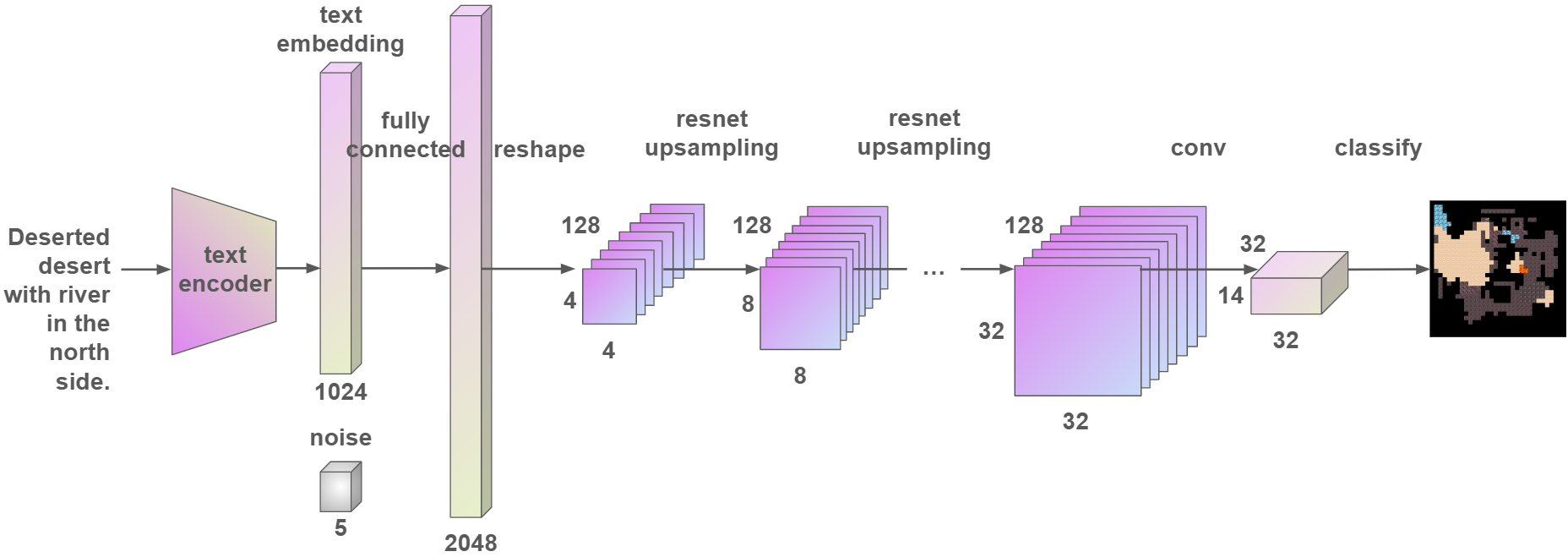}
    \caption{Five-Dollar Modelm architecture.}
    \label{fig:end-to-end-model}
\end{figure*}

\subsection{Five-Dollar Model (FDM)}
The Five-Dollar Model (FDM)m, proposed by~\citet{merino2023five}, is a streamlined model for Text-to-game-Map generation. It is a feed-forward neural network, which maps a text embedding vector directly to a map output. The transformation from text to map can be represented as:
\begin{equation*}
    M' = \mathcal{G}_{\theta}(\mathbf{T}).
\end{equation*}

The model is only designed to minimize the reconstruction error via self-supervised learning \revision{as:}
\begin{equation*}
    L = ||M - M'||_2^2.
\end{equation*}

In this strategy, training can be conceptualized as a classification task. The text embedding vector serves as a discriminative feature that the model uses to "classify" into the correct map representation. This perspective aligns the generative task with classification paradigms, focusing on feature utilization and output accuracy.

Fig.~\ref{fig:end-to-end-model} outlines the network architecture. Basic feedforward layers are employed to align the text embedding vector to the map. The text embedding vector is concatenated with a small noise vector, passed through a dense layer, and reshaped into a grid.  This is upsampling through three residual blocks, and finally reshaped to a $H \times W \times C$ map output via a convolutional layer.

\subsection{\revision{Discrete} Diffusion Model (DDM)}

To tackle the diversity challenge, we provide one specific method for training a Discrete Diffusion Model (DDM). It starts with a $H \times W \times C$ array of random normal noise \revision{map $m_{t}$} and gradually denoises it across multiple timesteps, each time moving closer to the map distribution. This iterative refinement uses a model that learns and predicts the noise to be removed at each step \cite{ogdiffusion}. Conditioning on the text embedding vector ensures that the denoising trajectory is aligned with the semantics of the input text, allowing the generation of maps that possess visual appeal and are contextually relevant to the input text. 

\begin{figure*}[h]
    \centering\includegraphics[width=0.85\textwidth]{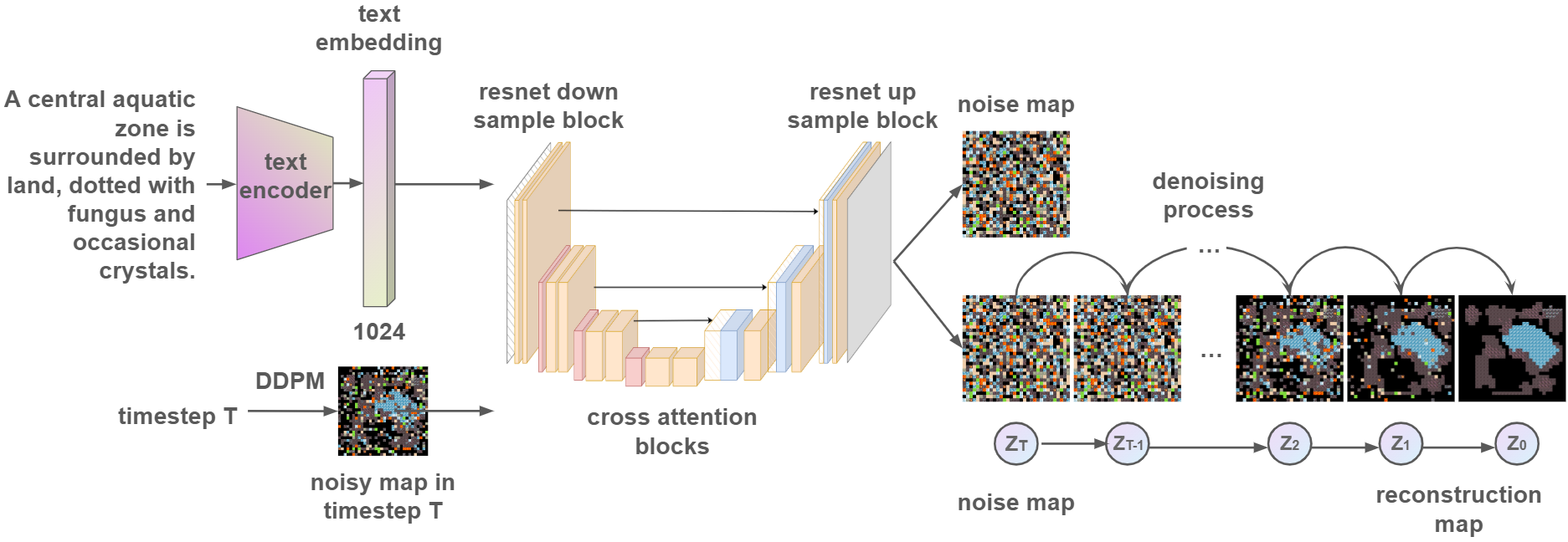}
    \caption{\revision{Discrete} Diffusion Model architecture.}
    \label{fig:discrete_ddpm}
\end{figure*}

As illustrated in Fig.~\ref{fig:discrete_ddpm}, our diffusion model $\mathcal{G}_{\theta}$ is based on a conditional UNet comprising Resnet and cross-attention blocks that effectively learn the constrained content passed from the text embeddings. At each time step $t$, the model takes a noisy map $m_t$ and the text embedding vector $\mathbf{t}$ as input, and outputs the predicted noise $\hat{\epsilon}$ of the map. We adopt the same setting as the Denoising Diffusion Probabilistic Model (DDPM)~\cite{ho2020denoising} to add the Gaussian noise to the map at each timestep. The denoising process can be mathematically formulated as follows:
\begin{equation*}
\begin{aligned}
m_t &= \text{DDPM}(\epsilon, m_0, t) = \sqrt {\bar{\alpha}_t} m_0 + \sqrt {1 - \bar{\alpha}_t}\epsilon, \\
&\text{where } \epsilon \sim \mathcal{N}(0, 1)  \text{ and } \hat{\epsilon} = \mathcal{G}_{\theta}(m_t, \mathbf{t}, t).
\end{aligned}
\end{equation*}

In the DDPM, $\bar{\alpha}_t$ denotes a decreasing scheduler. The training objective involves minimizing a loss function which focuses on the accurate prediction of noise, $\epsilon$. The loss function, $L$, is defined as follows: 
\begin{equation*}
    L = \mathbb{E}_{\mathbf{m}_0, \mathbf{t}, \mathbf{\epsilon}\sim \mathcal{N}(0,1), t\sim \mathcal{U}(1,T)}(\|\mathbf{\epsilon} - \mathcal{G}_{\theta}(\mathbf{m}_t, \mathbf{t}, t)\|_2^2).
\end{equation*}

This iterative denoising process continues until $t = 0$, resulting in a map $m_0$ that resembles the original training data while incorporating the semantics of input text $\mathbf{t}$. 

\section{Experiment and Result}
\subsection{Evaluation Methods}
We use OpenAI's GPT-4 Turbo (\textit{gpt-4-turbo-2024-04-09}) to generate map descriptions. Due to financial constraints, we only generate descriptions for 3,000 maps (in principle, an arbitrary number of descriptions could be generated). Each map is given 10 LLM-generated descriptions, divided into 5 long and 5 short descriptions. Long and short descriptions vary in detail and breadth. For example, the long and short descriptions for one map are:


\begin{itemize}
    \item \textbf{Long Description:}  \textit{A diverse terrain with four main areas, each featuring a combination of fungus and ground. The northwest region is dotted with stone and ashes amidst more ground and fungus.}
    \item \textbf{Short Description:} \textit{Four area division: ground, fungus, scarce stones, and ash fragments.}
\end{itemize}

\noindent{We evaluate the descriptions using the following metrics:}

\revision{
\begin{itemize}
    \item \textbf{BLEU} \cite{papineni2002bleu} and \textbf{ROUGE-L} \cite{lin2004rouge} scores assess the overlap between generated descriptions and human written references, indicating how closely the generated content resembles human language in terms of lexical choice and phrasing. Higher scores suggest that the model produces text that is more similar to human descriptions, reflecting better human-like quality.
    \item \textbf{METEOR} \cite{lavie2007meteor} considers synonymy and stem matching, providing insight into the semantic adequacy of the descriptions. Higher scores suggest that the content conveys the same meaning as human descriptions.
    \item \textbf{SPICE} \cite{anderson2016spice} focuses on the semantic propositional content, evaluating how well the generated descriptions capture the underlying meaning and relationships present in human descriptions. This metric is particularly relevant for assessing the richness of semantic content and the diversity of generated outputs.

    \item \textbf{CLIP Score} \cite{hessel2022clipscore}: Measures \textit{cross-modal alignment} between text and images using a pretrained vision-language model.
\end{itemize}

These metrics collectively assess the descriptions for lexical overlap, semantic content, and alignment with human-like language, ensuring diversity and contextual relevance.
}



\subsection{Evaluation of Generated Descriptions}
We separately evaluate the effects of the generated long and short descriptions, including comparisons within each type and against human references.\footnote{Additional comparisons can be found in the appendix.}

\subsubsection{Comparison within Generated Descriptions}
For each type of description, we selected the first description as a reference, calculated the scores against the other four descriptions, and averaged the results across all 3,000 maps (Table~\ref{tab:gpt4 long vs short}). Long descriptions outperformed short descriptions in almost all metrics, suggesting that long descriptions better align with human-like quality and capture more detailed and relevant information.


\begin{table}
\centering
\begin{tabular}{@{}clll@{}}
\toprule
\textbf{S.No} & \textbf{Metric} & \textbf{Des. Long} & \textbf{Des. Short} \\ \midrule
\textbf{0} & Bleu\_1  & 54.71 & \textbf{58.37} \\
\textbf{1} & Bleu\_2  & \textbf{26.57} & 26.56 \\
\textbf{2} & Bleu\_3  & \textbf{12.18} & 11.68  \\
\textbf{3} & Bleu\_4  & \textbf{05.62}  & 04.86   \\
\textbf{4} & METEOR   & \textbf{19.50} & 12.64 \\
\textbf{5} & ROUGE\_L & \textbf{33.27} & 26.16 \\
\textbf{6} & SPICE    & \textbf{11.31} & 03.89  \\ \bottomrule
\end{tabular}
\caption{Comparison within generated descriptions.}
\label{tab:gpt4 long vs short}
\end{table}

\subsubsection{Generated Descriptions versus Human References}
We compared synthetic captions with 30 randomly selected human references to evaluate how closely synthetic captions resemble human descriptions (Table~\ref{tab:gpt4_H long vs short}). These results are similar to the previous comparison. Long descriptions more effectively mimic the human descriptive style than short descriptions, likely because they incorporate more detail, making them appear more natural and comprehensive.

The use of random human references as a benchmark is insightful, revealing that long synthetic descriptions maintain strong linguistic and semantic quality, even when compared against diverse human-written descriptions.


\begin{table}
\centering
\begin{tabular}{@{}clll@{}}
\toprule
\textbf{S.No} & \textbf{Metric} & \textbf{Des. Long} & \textbf{Des. Short} \\ \midrule
\textbf{0} & Bleu\_1  & \textbf{70.84} & 50.58 \\
\textbf{1} & Bleu\_2  & \textbf{43.47} & 21.13 \\
\textbf{2} & Bleu\_3  & \textbf{23.56} & 07.88 \\
\textbf{3} & Bleu\_4  & \textbf{11.49} & 02.38 \\
\textbf{4} & METEOR   & \textbf{16.56} & 09.38 \\
\textbf{5} & ROUGE\_L & \textbf{42.92} & 23.25 \\
\textbf{6} & SPICE    & \textbf{03.43} & 00.77 \\ \bottomrule\\
\end{tabular}
\caption{Comparison between generated descriptions and human references.}
\label{tab:gpt4_H long vs short}
\end{table}

\begin{table}[h]
    \centering
    \begin{tabular}{ccc}
        \toprule
        \textbf{Brogue} & \textbf{FDM} & \textbf{DDM} \\
        \toprule
        \includegraphics[width=0.12\textwidth]{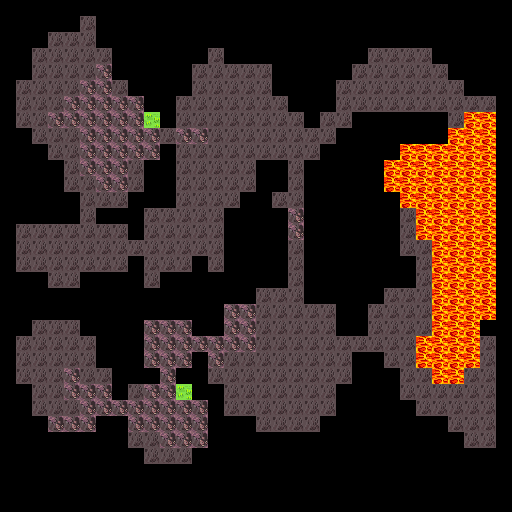} &
        \includegraphics[width=0.12\textwidth]{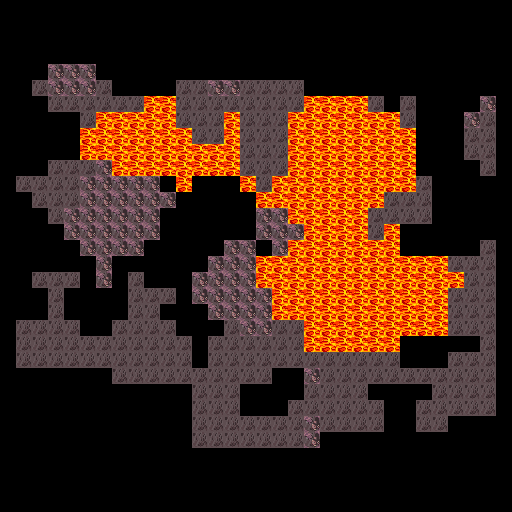} &
        \includegraphics[width=0.12\textwidth]{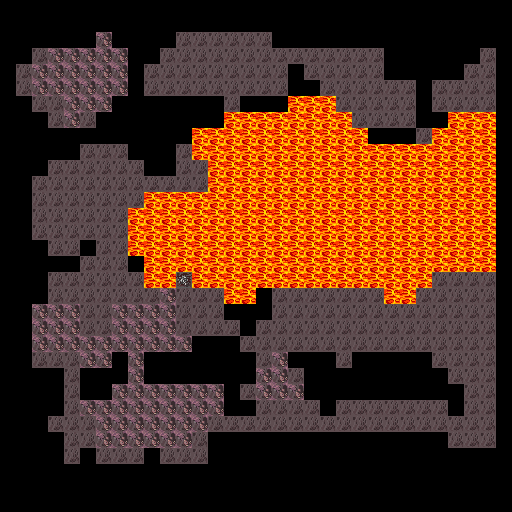} \\
        \hline
        \includegraphics[width=0.12\textwidth]{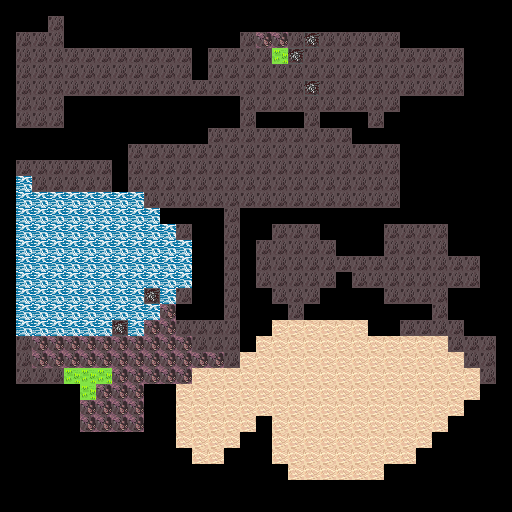} &
        \includegraphics[width=0.12\textwidth]{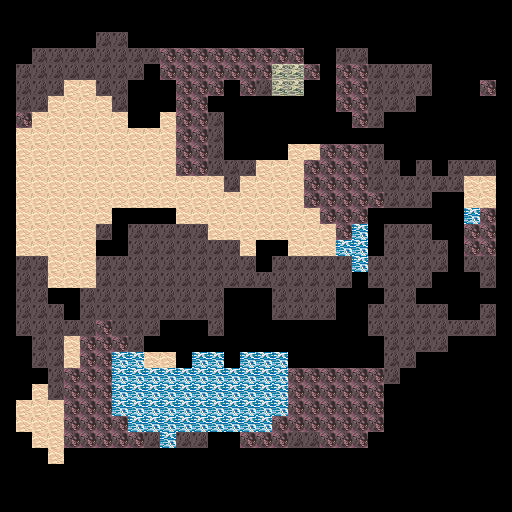} &
        \includegraphics[width=0.12\textwidth]{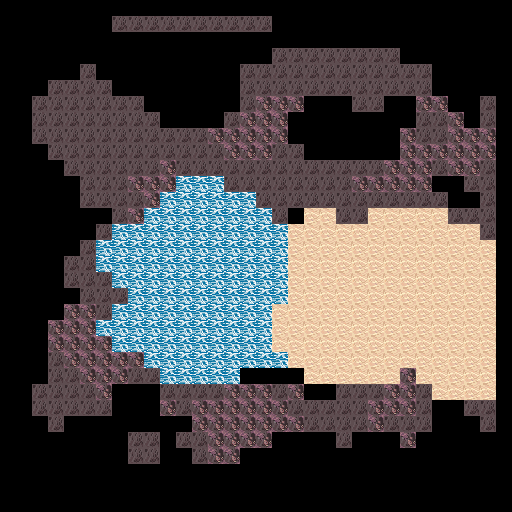} \\
        \hline
        \includegraphics[width=0.12\textwidth]{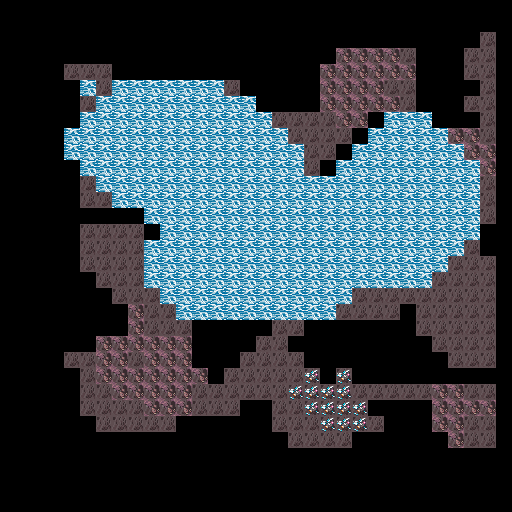} &
        \includegraphics[width=0.12\textwidth]{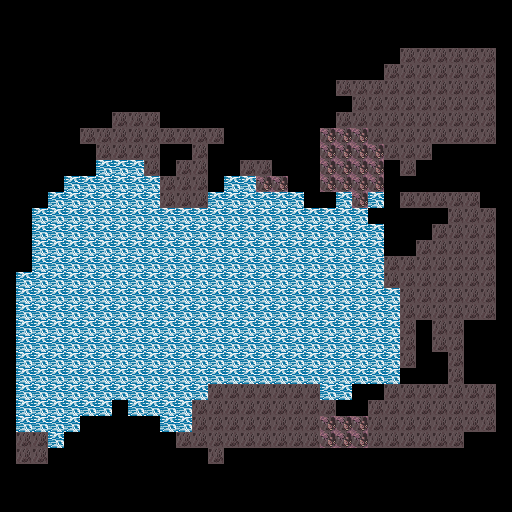} &
        \includegraphics[width=0.12\textwidth]{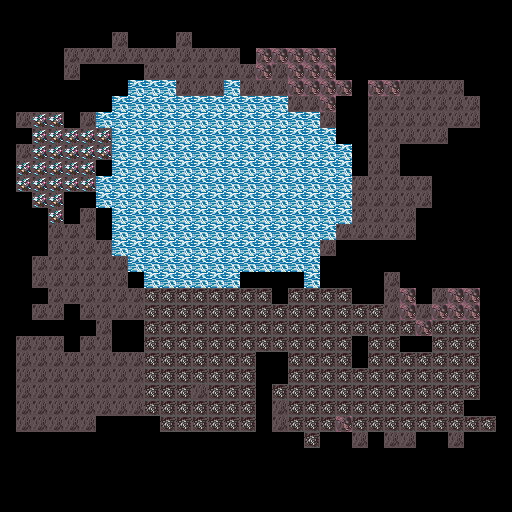} \\
        \hline
        \includegraphics[width=0.12\textwidth]{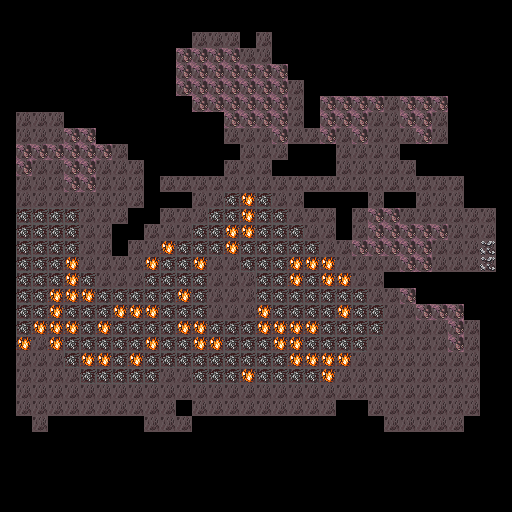} &
        \includegraphics[width=0.12\textwidth]{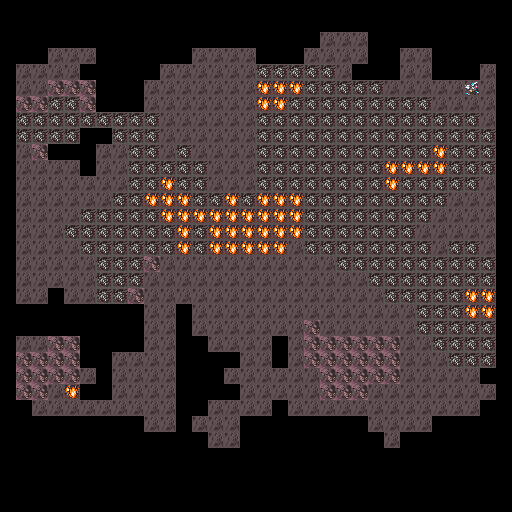} &
        \includegraphics[width=0.12\textwidth]{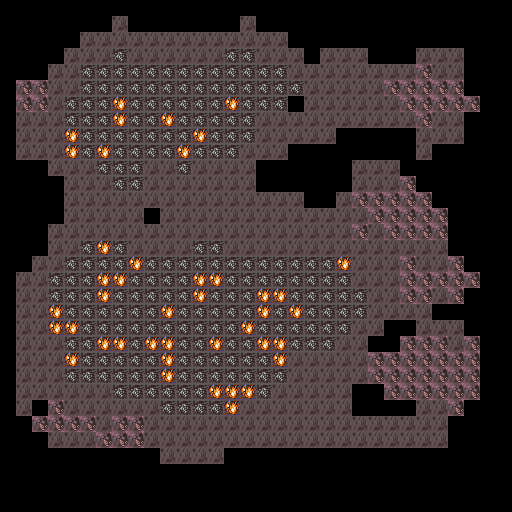} \\
        \toprule
    \end{tabular}
    \caption{Generated maps based on generated  descriptions: (1) A \textcolor{orange}{central} area \textcolor{blue}{dominated} by \textcolor{blue}{a large expanse} of \textcolor{red}{lava} surrounded by \textcolor{red}{solid ground}, with \textcolor{blue}{sparsely} growing \textcolor{red}{fungi} and \textcolor{blue}{a few patches of} \textcolor{red}{grass}; (2) A \textcolor{blue}{vast} \textcolor{red}{desert} landscape merges with clear \textcolor{red}{blue waters}, spotted with \textcolor{red}{fungus} growths and \textcolor{blue}{small} patches of \textcolor{blue}{greenery} amidst the \textcolor{red}{sandy terrain}; (3) A \textcolor{orange}{central} \textcolor{red}{aquatic zone} is surrounded by \textcolor{red}{land}, \textcolor{blue}{dotted} with \textcolor{red}{fungus} and \textcolor{blue}{occasional} crystals. Two smaller regions lie on the \textcolor{orange}{eastern edge}, connected by pathways to the main area; (4) A \textcolor{red}{charred landscape} with \textcolor{red}{scorched earth} and \textcolor{red}{rampant wildfires} spread across. \textcolor{blue}{Patches} of \textcolor{red}{fungi} cling to life amidst the destruction, with rare \textcolor{red}{mineral-rich stones} scattered \textcolor{blue}{sparsely}.}
    \label{table:generation1}
\end{table}

\begin{table}[h]
    \centering
    \begin{tabular}{ccc}
        \toprule
        \textbf{FDM} & \textbf{DDM-1} & \textbf{DDM-2} \\
        \toprule
        \includegraphics[width=0.12\textwidth]{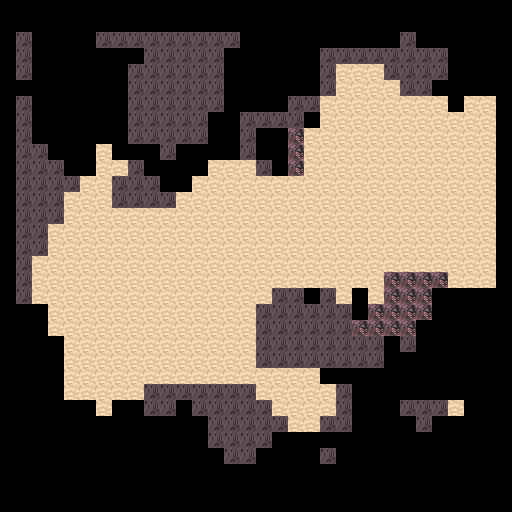} &
        \includegraphics[width=0.12\textwidth]{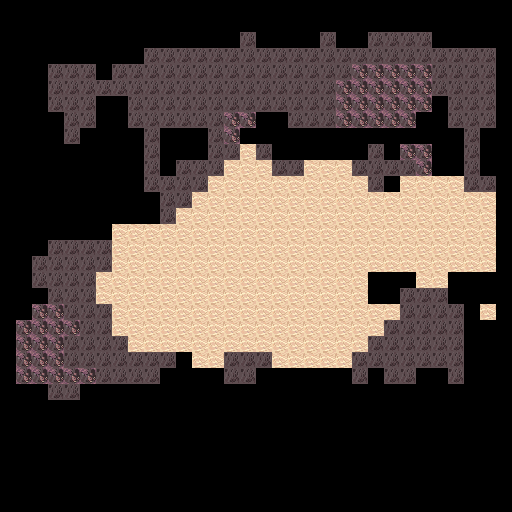} &
        \includegraphics[width=0.12\textwidth]{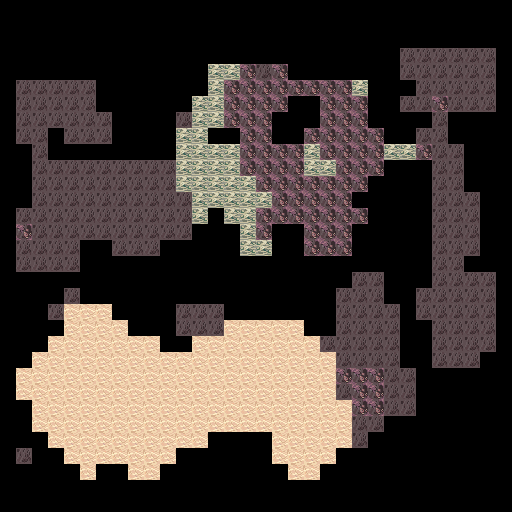} \\
        \hline
        \includegraphics[width=0.12\textwidth]{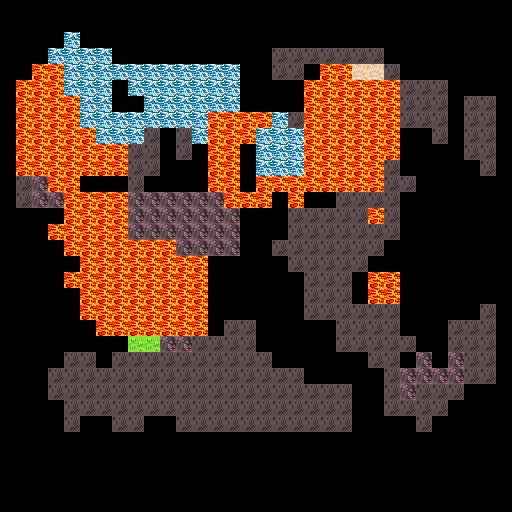} &
        \includegraphics[width=0.12\textwidth]{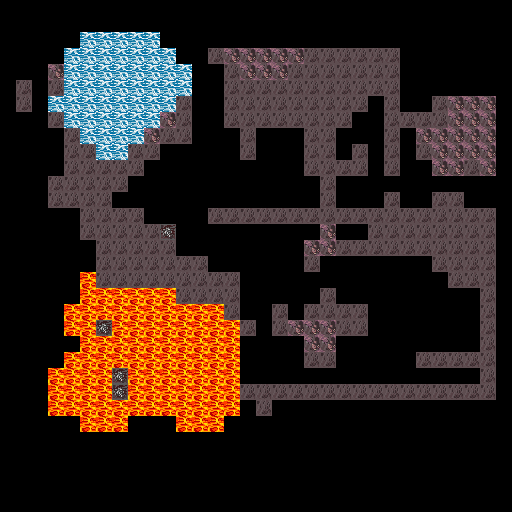} &
        \includegraphics[width=0.12\textwidth]{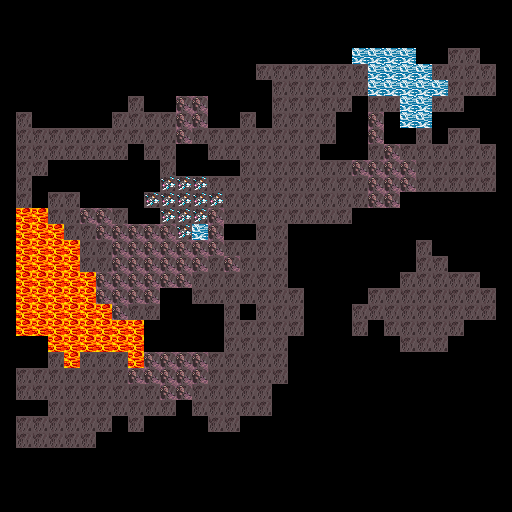} \\
        \hline
        \includegraphics[width=0.12\textwidth]{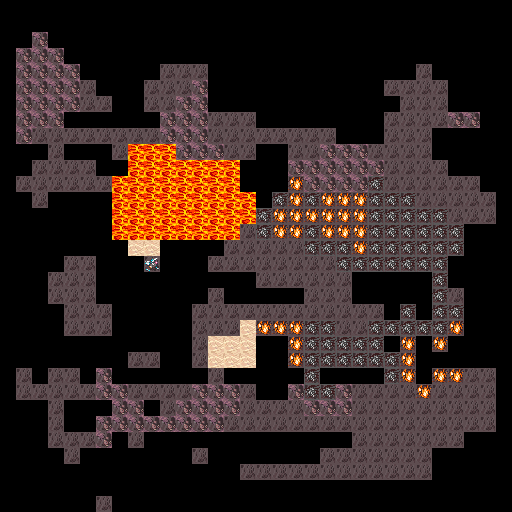} &
        \includegraphics[width=0.12\textwidth]{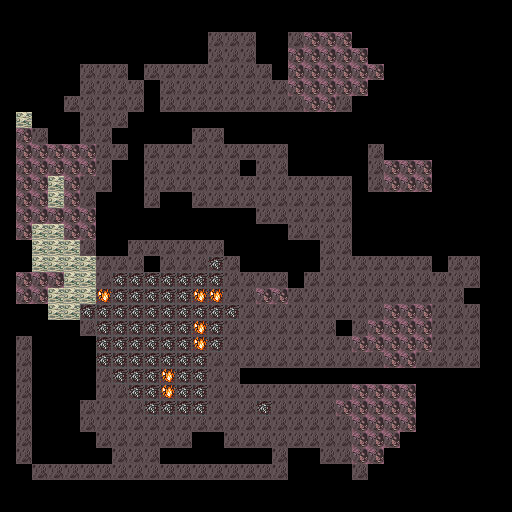} &
        \includegraphics[width=0.12\textwidth]{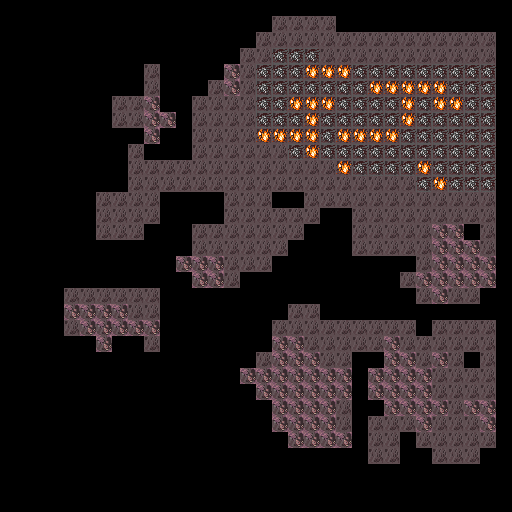} \\
        \hline
        \includegraphics[width=0.12\textwidth]{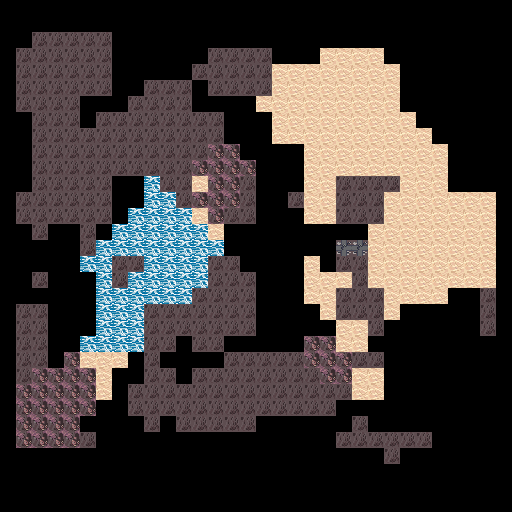} &
        \includegraphics[width=0.12\textwidth]{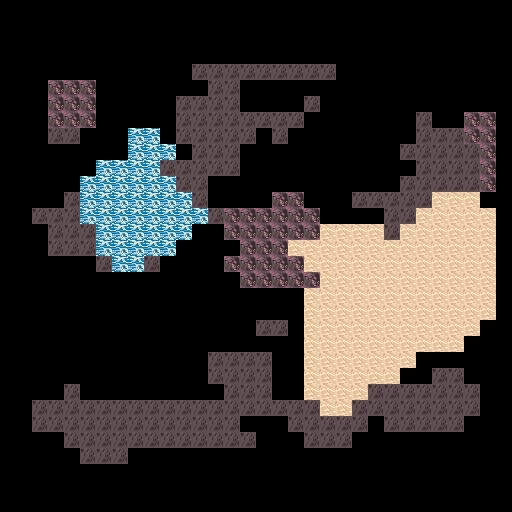} &
        \includegraphics[width=0.12\textwidth]{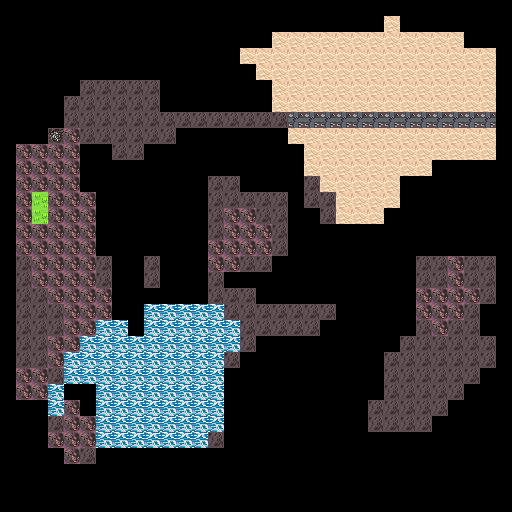} \\
        \toprule
    \end{tabular}
    \caption{Generated maps based on human prompts: (1) A \textcolor{blue}{vast} \textcolor{red}{sandy} area; (2) Some \textcolor{red}{lakes} in the \textcolor{orange}{north}, and \textcolor{blue}{a lot of} \textcolor{red}{magma and lava}; (3) \textcolor{red}{Burning out fire} in the \textcolor{orange}{center}; (4) The \textcolor{red}{lake} to the \textcolor{orange}{left}. The \textcolor{red}{desert} to the \textcolor{orange}{right}. Connected by \textcolor{red}{bridges}.}
    \label{table:generation2}
\end{table}

\subsection{Evaluation of Generated Maps}
Tables~\ref{table:generation1} and \ref{table:generation2} show generated maps using the Moonshine distillation process. Table~\ref{table:generation1} compares source maps from Brogue with generated maps, using LLM-generated descriptions. Table~\ref{table:generation2} compares source and generated maps using human-authored input. Specific tile types, locations, and ranges in prompts are highlighted in  \textcolor{red}{red},  \textcolor{blue}{blue}, and  \textcolor{orange}{orange}, respectively. We see that DDM can produce a diverse set of maps compared to FDM. However, under a close evaluation FDM sometimes captures nuances of the text slightly better. The FDM sand tiles cover nearly the entire map in "A vast sandy area", whereas the sand is a smaller portion in the DDM generations. When asked to create "some lakes", FDM produces multiple lakes whereas DDM produces only one lake. As the diffusion method relies on data, it is possible that with more examples given, the model may start to understand these smaller nuances in the prompts.

\paragraph{Quantitative Evaluation}
The models demonstrate reasonable accuracy in reflecting specific types of tiles mentioned in the descriptions, especially for the main area. Both models excel at capturing descriptions like \textit{"central area"} or \textit{"a vast area of..."} by generating wide regions that include the specified tiles. However, they sometimes struggle with more detailed elements, such as \textit{"a dot of..."}, \textit{"scattered around,"} or \textit{" fews..."} This limitation may stem from synthetic descriptions that primarily focus on the main area while neglecting finer details. 

While not all generated regions are connected, the models exhibit an understanding of hidden \textit{connectivity} and \textit{ecology} features beyond explicit descriptions. FDM struggles with diversity, producing fixed outputs based on a given prompt due to its focus on learning mapping relationships. In contrast, DDM outperforms it by generating \textit{varied} results that align with the text and include additional details.


\begin{figure}
    \centering
    \includegraphics[width=0.3\textwidth]{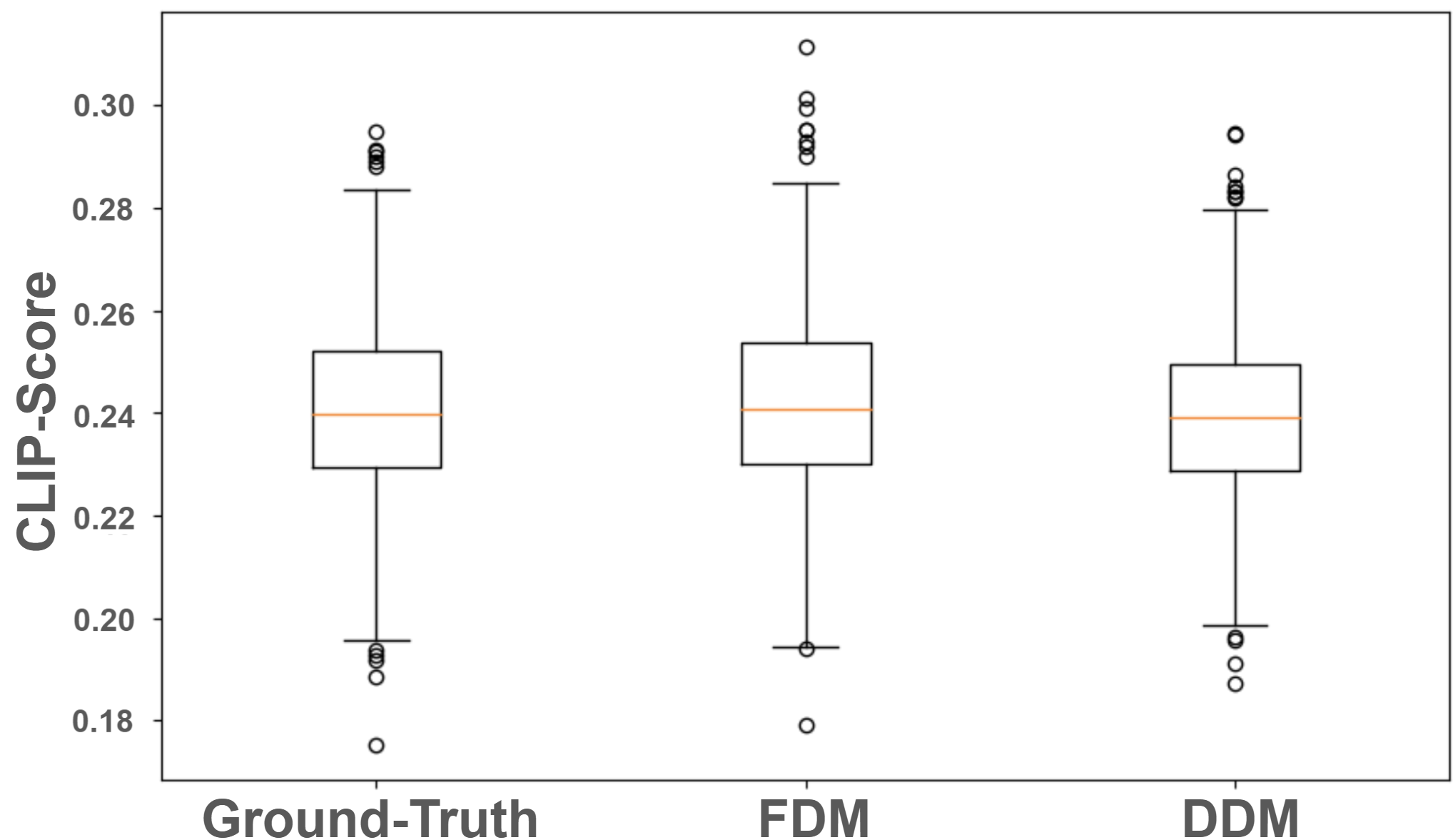}
    \caption{CLIP score of FDM (left) and DDM (right) v.s. the Brogue ground truth (X-axis).}
    \label{fig:pt_scatter_compare}
\end{figure}

\paragraph{\revision{Similarity between Generated Map and Ground Truth}} We calculate the CLIP Score between the ground truth maps Brogue generated and the maps generated by the FDM or DDM, respectively. The scatter plots are shown in Fig.~\ref{fig:pt_scatter_compare} which exhibits a concentrated distribution of data points, showing a strong correlation of the similarity between the Brogue ground truth and the two generative models’ predictions. DDM demonstrates a high degree of clustering, indicating good predictive accuracy. 

\begin{figure}[h]
    \centering
    \includegraphics[width=0.47\textwidth]{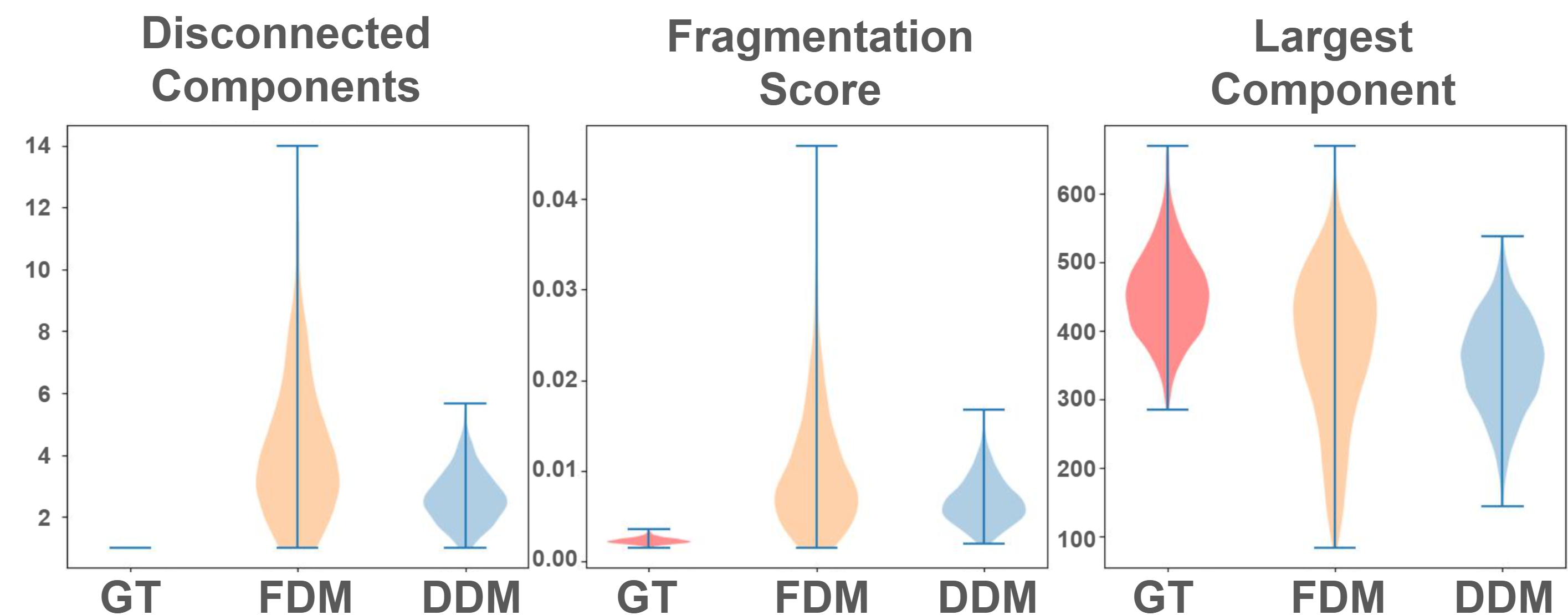}
    \caption{Connectivity analysis of models based on three metrics: the number of disconnected components, fragmentation score, and largest component size. DDM has fewer disconnected components, lower fragmentation, and more stable component sizes compared to FDM.}
    \label{fig:connectivity}
\end{figure}

\revision{  
To compare the diversity of DDM and the FDM, we perform connected component analyses to assess usable structural diversity. The results in Fig. \ref{fig:connectivity} demonstrate that the DDM generates maps with a wider range of usable patterns. }

\section{Conclusion}

We introduce Moonshine, a method for distilling traditional PCGML algorithms into text-conditioned PGCML models using synthetic data generation. Based on an automated map-labeling approach via Large Language Models (LLM), we minimize the labor for human annotation --- a major obstacle for Text-to-game-Map (T2M) models. Additionally, we provide our open-source dataset of dungeon maps with associated descriptions for use in further T2M research. 

By combining traditional constructive PCG algorithms with the semantic understanding of LLMs, we enable the creation of diverse, human-like synthetic descriptions for existing PCG map data. We found that longer descriptions generated by LLMs capture more semantic information and align more closely with human references, whereas shorter descriptions correspond more with map image information in pretrained CLIP models. However, our fine-tuned CLIP model shows that longer descriptions integrate better with maps, suggesting that while a frozen model may perform better with short descriptions, longer descriptions are preferable for fine-tuning multimodal-aligned models.

We explored two T2M models —the Five-Dollar Model (FDM) and the Discrete Diffusion Model (DDM) —— and found that while both generate maps correlated with their respective text descriptions, longer descriptions offer more accurate semantic reflections. FDM utilizes a feed-forward network to directly map text embeddings onto a map, achieving slightly higher overall scores compared to the DDM. However, it lacked diversity, producing relatively uniform results. In contrast, the DDM, which incrementally restores a noise map to a reconstructed map, shows greater potential for diversity. 

However, the limitations in the dataset size for training the DDM suggest that extending training over more epochs with a larger synthetic dataset could improve its performance. \revision{We acknowledge that a loss of fine-grained control is a limitation of our current model, partly due to constraints in LLM annotations and the available training data. We anticipate that expanding the dataset to include additional tile types could mitigate this issue and enhance the model’s level of control. Additionally,} A larger dataset of human-written descriptions would provide valuable information regarding the human likedness of the generated descriptions. A study should be conducted to assess the use of the distilled Moonshine models with a designer trying to construct specific map outputs. We look only at a subset of procedural dungeon generation in terrain: items, enemies, and even stories all contribute to better model distillation.

In summary, Moonshine's automated distillation generation process can be an effective method for creating controllable content generators. For optimal results in description generation and model selection, we recommend prioritizing longer descriptions and considering the Discrete Diffusion Model for tasks requiring diverse outputs. Future research can explore ways to encode additional metadata into the map descriptions, such as items, enemies, and stories, or incorporate reinforcement learning (RL) mechanisms to further refine control over generated maps, allowing dynamic user interaction during generation.

\appendix
\section{\textsc{Appendix}}




\subsection{Supplemental Experiments}

\subsubsection{Pre-trained v.s. Fine-tuned CLIP Model for Synthetic Descriptions}
In the histogram for the pre-trained and fine-tuned CLIP model shown in Fig. \ref{fig:pt_gpt_l_s_} and \ref{fig:ft_gpt_l_s_} respectively, the distribution reveals a notable improvement in the performance of long descriptions (blue) in the fine-tuned model compared to short ones (orange). The long descriptions have a larger maximum and a smaller variance. This suggests that the fine-tuning process allows the model to better align detailed, complex descriptions with image content, consistently outperforming shorter descriptions.

\subsubsection{Diversity of the Synthetic Descriptions}
We also evaluated the expressive range of different synthetic descriptions by running multiple iterations with the same prompt and comparing the edit distance~\citep{lcvenshtcin1966binary} of the outputs. Fig. \ref{fig:edit_distance} shows a similar average edit distance of 500 steps between the same pre-generation prompt indicating a high degree of diversity.

\subsubsection{Visualization of DDM}

\begin{figure}[H]
     \begin{minipage}{0.5\textwidth}
     \centering
    \includegraphics[width=0.65\textwidth]{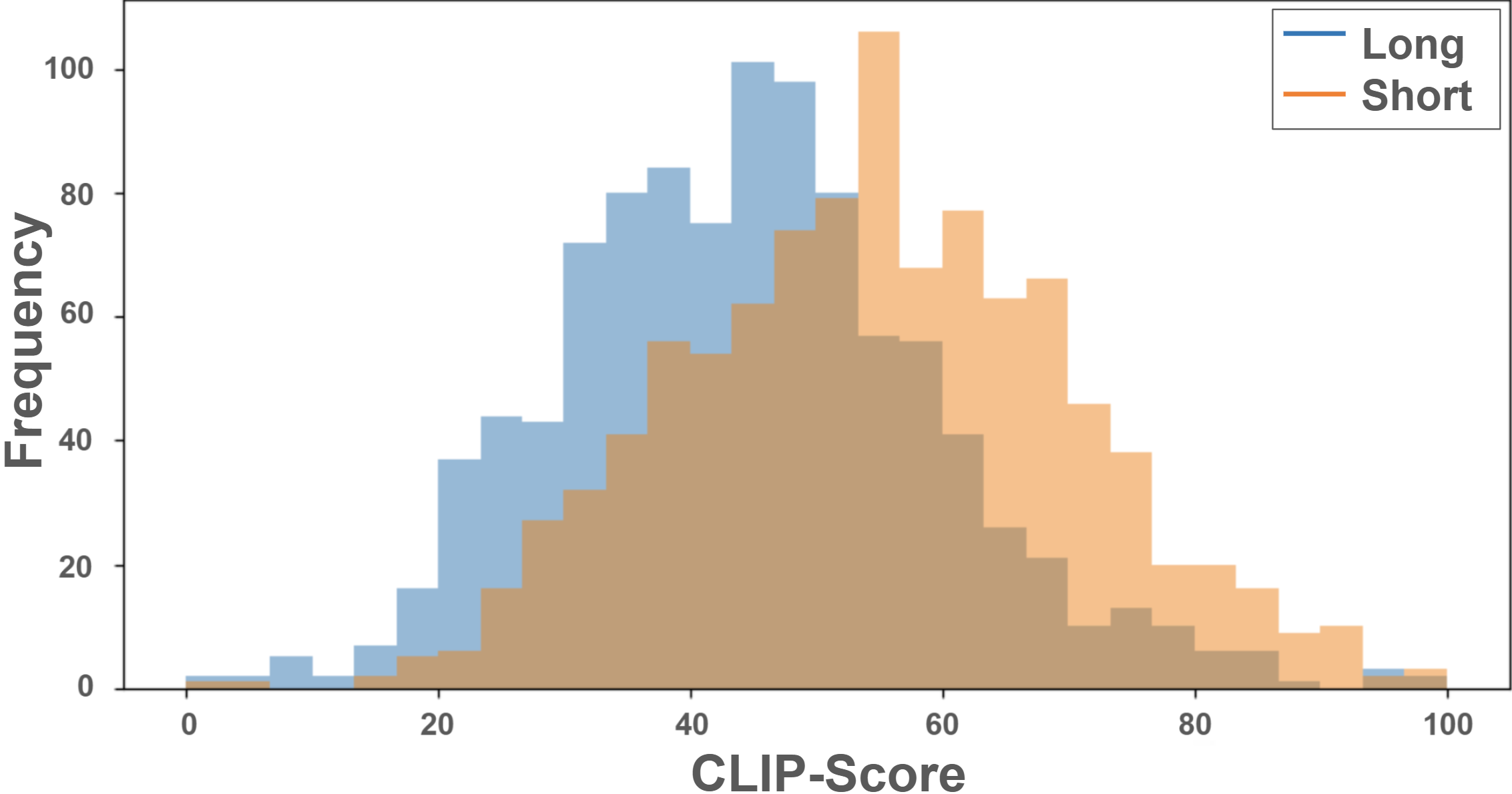}
    \caption{Normalized pre-trained CLIP Score for Long (blue) v.s. Short Descriptions (orange).}
    \label{fig:pt_gpt_l_s_}
    \centering
    \includegraphics[width=0.65\textwidth]{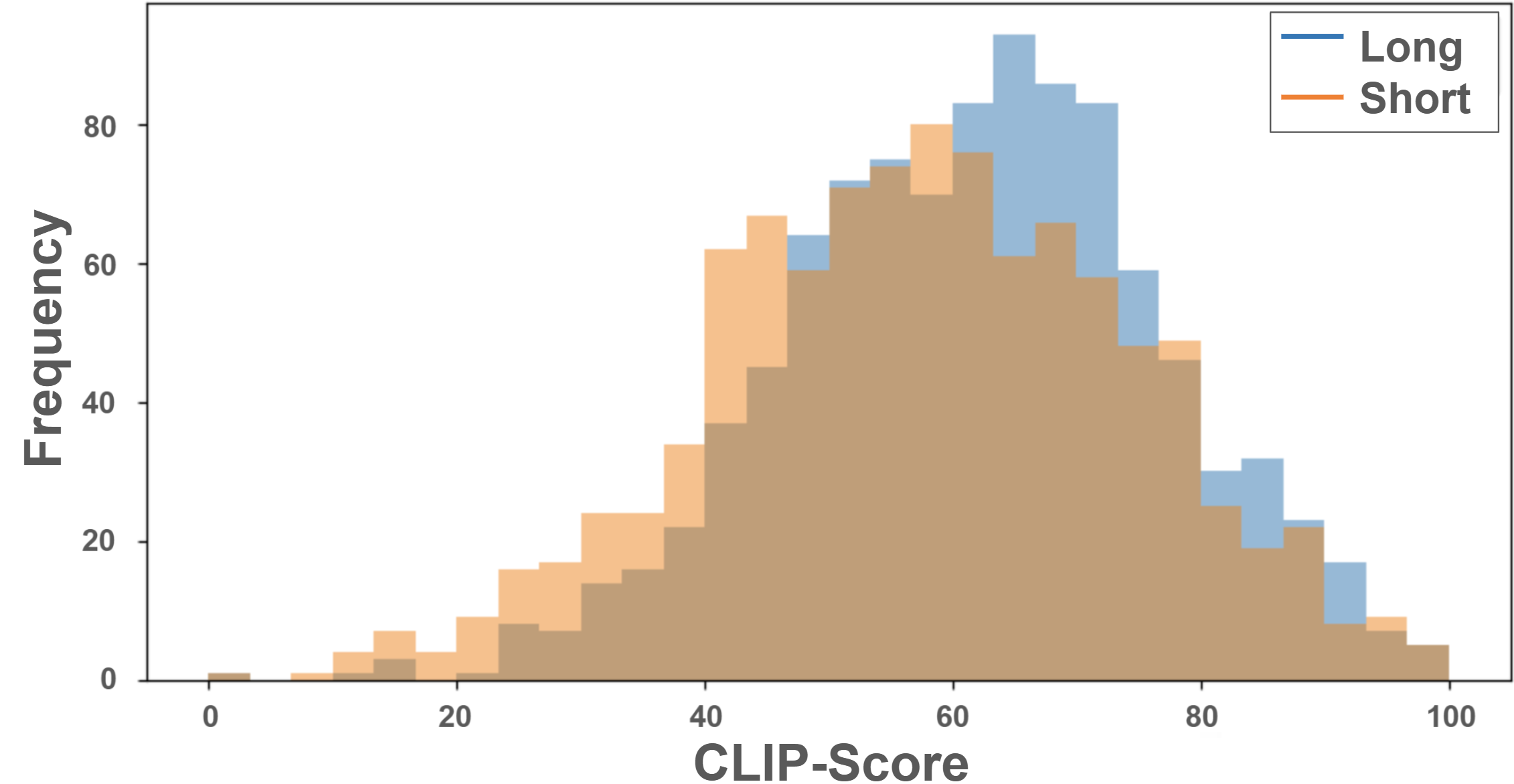}
    \caption{Normalized fine-tuned CLIP Score for Long (blue) v.s. Short Descriptions (orange).}
    \label{fig:ft_gpt_l_s_}
    \centering
    \includegraphics[width=0.65\textwidth]{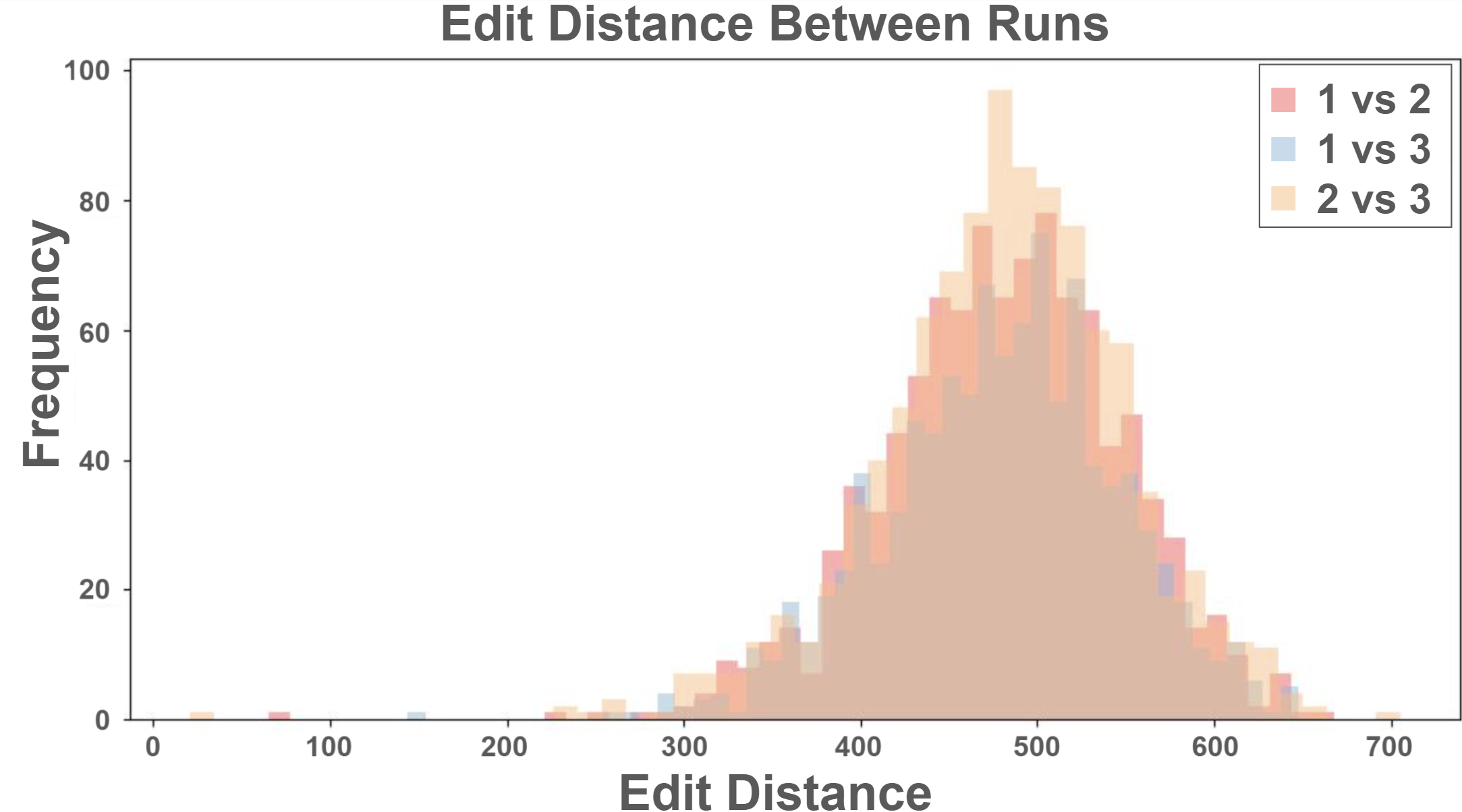}
    \caption{Edit distance between different outputs based on the same prompt generated by the LLM.}
    \label{fig:edit_distance}
    \centering
    \includegraphics[width=0.6\textwidth]{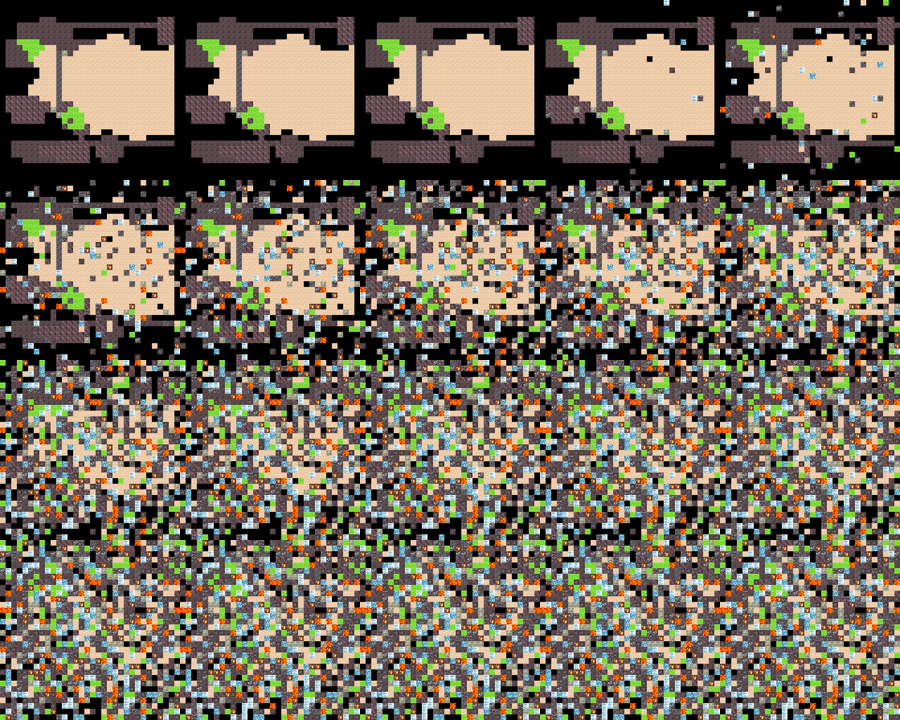}
    \caption{Visualization of DDM across different timestamps.}
    \label{fig:ddm_visual}
    \end{minipage}
\end{figure}

As shown in Fig. \ref{fig:ddm_visual}, We output the probability distribution of the discrete diffusion model over a certain timestamp and visualize DDM by selecting the cell with the largest probability distribution. It is observed that DDM has similar characteristics to the general Diffusion model.

\

\subsubsection{Reconstruction Loss} 
\begin{figure}[h]
    \centering
    \begin{subfigure}[b]{0.35\textwidth}
        \centering
        \includegraphics[width=\textwidth]{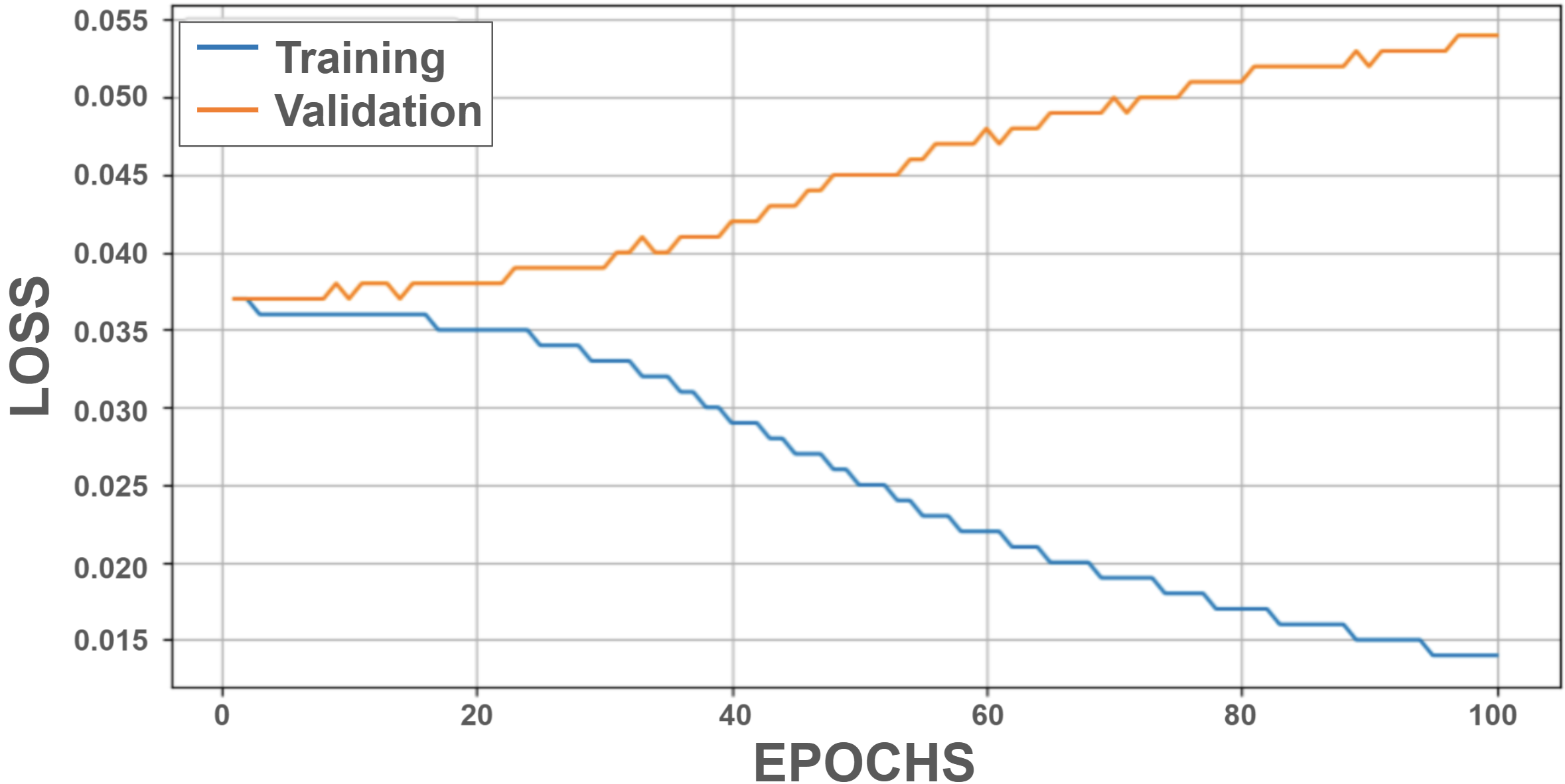}
        \caption{FDM}
        \label{fig:5dollar_test_train}
    \end{subfigure}
    \begin{subfigure}[b]{0.35\textwidth}
        \centering
        \includegraphics[width=\textwidth]{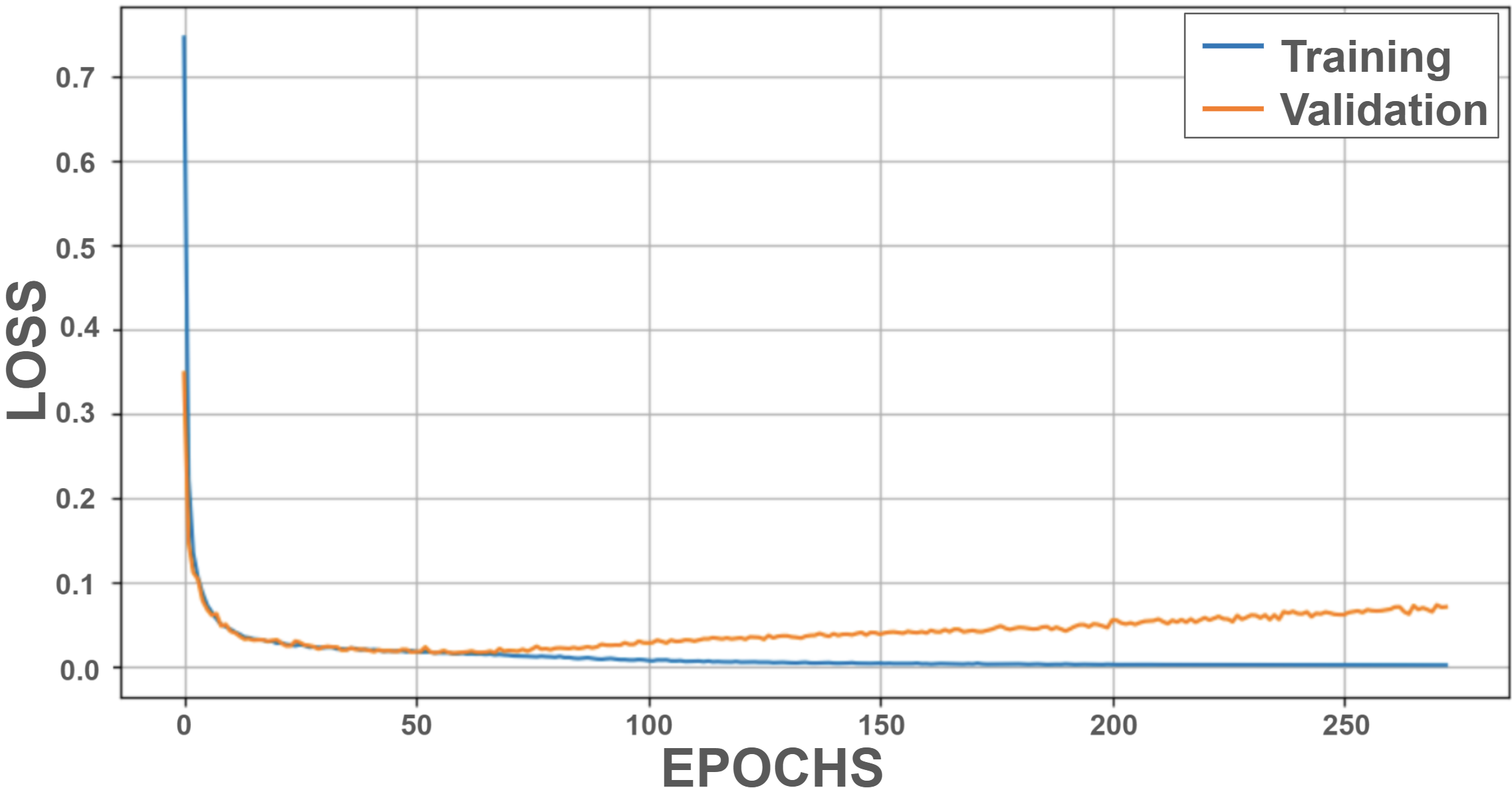}
        \caption{DDM}
        \label{fig:diffusion_test_train}
    \end{subfigure}
    \caption{Reconstruction loss (MSE) during training (blue) and validation (orange) phases.}
    \label{fig:test_train}
\end{figure}

We compared the reconstruction loss (Mean Squared Error, MSE) of two models during the training and generation phases, as shown in Fig. \ref{fig:5dollar_test_train} and Fig. \ref{fig:diffusion_test_train}. FDM achieves a validation loss of 0.037–0.054 but overfits after 20 epochs, learning only the training dataset's mappings. In contrast, DDM converges within 20 epochs but shows a slower increase in validation loss even after 250 epochs, indicating better generalization and the ability to generate diverse unseen data while matching FDM's performance.

\subsection{Prompt Formation\footnote{Due to space constraints, the full prompt is available in our Hugging Face repository.}}

\subsubsection{Pre-generation Prompt}
Our prompt is divided into four sections: \textbf{Setting}, \textbf{Response Format}, \textbf{Examples}, and \textbf{Rules}, using a markdown format with highlighted keywords as described in \cite{he2024automated}. 

\textbf{(1) Setting:} The LLM is assigned as a \textit{helpful data annotator} capable of generating sentences beyond standard English grammar. It processes a \textit{2D text-to-game-map generative model} with a given \textit{map} and \textit{metadata} where integers correspond to tiles. \textbf{(2) Response Format:} The LLM generates \textit{10 diverse, human-like, and creative descriptions} of the map, ensuring \textit{no repetitions}. \textbf{(3) Examples:} Few-shot examples of human-authored map descriptions are provided, demonstrating the desired diversity and format based on metadata. \textbf{(4) Rules:} The LLM is instructed to balance descriptions between \textit{1–3 sentences} and \textit{5–15 words}, avoid repetitive terms like "serene" or "labyrinth," describe all major map areas, and ensure no repeated descriptions or pronouns in metadata.

\subsubsection{Generation Round}
During each generation, we provide the following information: \textbf{(1) Detailed Setting:} The model is designated as a \textit{helpful data annotator} tasked with processing a \textit{2D text-to-game-map generative model}.  \textbf{(2) Tile Information:} The model receives basic information about the tiles included in a map and their significance. \textbf{(3) Map and Metadata:} The LLM is provided with a character array and metadata that define the map, where integers correspond to specific tile types (e.g., \texttt{G\_GROUND}, \texttt{G\_WATER}). Metadata includes main and sub-areas, directional annotations (e.g., \texttt{N}, \texttt{SW}), and additional details about the areas’ tile composition and spatial ranges. \textbf{(4) Rules:} The model generates \textit{ten diverse, human-like descriptions} of 10–150 words each, ensuring no repetitions. It must describe all major areas while adhering to specific constraints, such as avoiding repetitive words, balancing short and long descriptions, and strictly following the metadata format.

\bibliography{biblography.bib}
\end{document}